\documentclass{article}

\usepackage{microtype}
\usepackage{graphicx}
\usepackage{xcolor}
\usepackage{subcaption}
\usepackage{booktabs}
\usepackage{caption}
\usepackage{multirow}

\usepackage{hyperref}
\usepackage{float}  
\usepackage{stfloats}

\usepackage[preprint]{icml2026}

\usepackage{amsmath}
\usepackage{amssymb}
\usepackage{mathtools}
\usepackage{amsthm}

\usepackage{wasysym}

\newcommand{\XSolidBrush}{\boldsymbol{\times}}    
\newcommand{\CheckmarkBold}{\boldsymbol{\checkmark}}

\usepackage[capitalize,noabbrev]{cleveref}

\theoremstyle{plain}

\theoremstyle{definition}

\theoremstyle{remark}

\usepackage[textsize=tiny]{todonotes}

\icmltitlerunning{FACTOR: Counterfactual Training-Free Test-Time Adaptation for Open-Vocabulary Object Detection}

\begin{document}

\twocolumn[
  \icmltitle{FACTOR: Counterfactual Training-Free Test-Time Adaptation for Open-Vocabulary Object Detection}



 \icmlsetsymbol{equal}{*}

 \begin{icmlauthorlist}
 \icmlauthor{Kaixiang Zhao}{yyy}
 \icmlauthor{Mao Ye}{yyy}
 \icmlauthor{Lihua Zhou}{yyy}
 \icmlauthor{Hu Wang}{yyy}

  \icmlauthor{Luping Ji}{yyy}
  \icmlauthor{Song Tang}{xxx}
  \icmlauthor{Xiatian Zhu}{zzz}
  \end{icmlauthorlist}

  \icmlaffiliation{yyy}{School of Computer Science and Engineering, University of Electronic Science and Technology of China, Chengdu, China}
  \icmlaffiliation{xxx}{Institute of Machine Intelligence (IMI), University
of Shanghai for Science and Technology, Shanghai, China}
  \icmlaffiliation{zzz}{Surrey Institute for People-Centred Artificial Intelligence, CVSSP, University of Surrey, Guildford, UK}

  \icmlcorrespondingauthor{Mao Ye}{cvlab.uestc@gmail.com}

  \icmlkeywords{Machine Learning, ICML}

  \vskip 0.3in
]



\printAffiliationsAndNotice{}  

\begin{abstract}
Open-vocabulary object detection often fails under distribution shifts, as it can be misled by spurious correlations between non-causal visual attributes (e.g., brightness, texture) and object categories. Existing test-time adaptation (TTA) methods either depend on costly online optimization or perform global calibration, overlooking the attribute-specific nature of these failures. To address this, we propose {\bf FACTOR} (counterFACtual training-free Test-time adaptation for Open-vocabulaRy object detection), a lightweight framework grounded in counterfactual reasoning. By perturbing test images along non-causal attributes and comparing region-level predictions between original and counterfactual views, FACTOR quantifies attribute sensitivity, semantic relevance, and prediction variation to selectively suppress attribute-dependent predictions—without parameter updates. Experiments on PASCAL-C, COCO-C, and FoggyCityscapes show that FACTOR consistently outperforms prior TTA methods, demonstrating that explicit counterfactual reasoning effectively improves robustness under distribution shifts. {\it Code will be released upon acceptance}.
\end{abstract}

\section{Introduction} \label{submission}
Open-vocabulary object detectors (OVOD) based on vision–language models (VLMs), such as Grounding DINO~\cite{liu2024grounding}, have demonstrated strong generalization ability by leveraging language-level semantic priors. However, when deployed in real-world environments, these detectors still suffer from significant performance degradation under distribution shifts caused by adverse weather, sensor noise, or synthetic corruptions. In many practical scenarios, collecting target-domain annotations or retraining the model is infeasible, making test-time adaptation (TTA) a particularly appealing solution for robust deployment. Recent TTA methods can be broadly categorized into training-based~\cite{zhang2024historical, gao2025testtimeadaptiveobjectdetection} and training-free paradigms~\cite{karmanov2024efficient, zhou2025bayesian}. Training-based approaches adapt model parameters online via self-supervision or entropy minimization, often requiring backpropagation and careful hyper-parameter tuning. In contrast, training-free TTA methods aim to adjust model predictions directly at inference time, offering better invariance and efficiency.

\begin{figure}[t]
\begin{center}
\centerline{\includegraphics[width=\columnwidth]{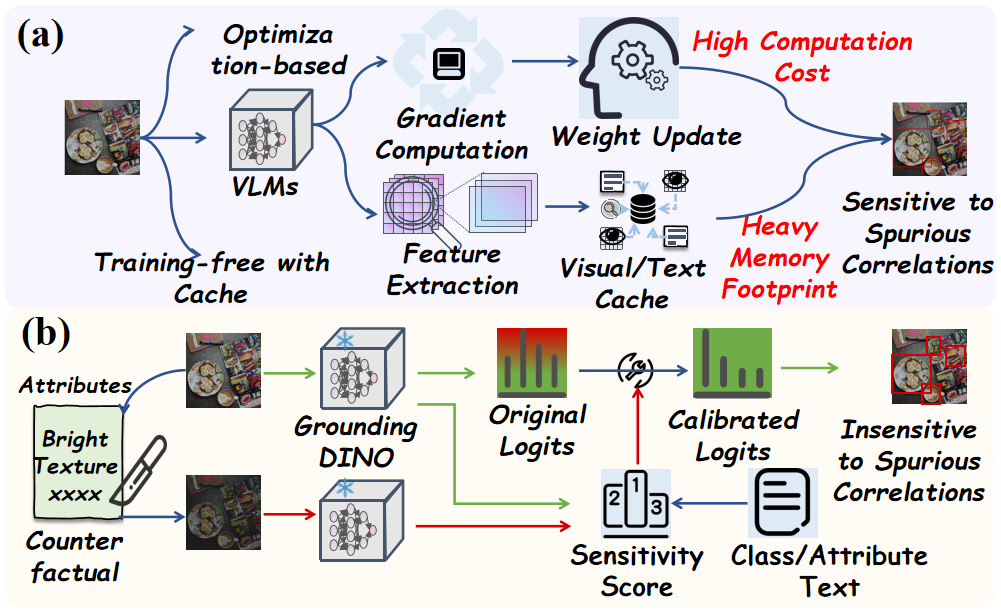}}
\caption{Comparison between current TTA approaches and FACTOR. (a) Previous methods either rely on costly on-line optimization or global calibration, overlooking fine-grained non-causal attribute interference. (b) FACTOR identifies and suppresses spurious attribute correlations by constructing counterfactual sample, efficiently refining predictions without parameter updating.}
\label{fig1}
\end{center}
\vskip -0.4in
\end{figure}

Despite recent advances in test-time adaptation (TTA), most methods operate at a global level, relying on aggregated confidence scores or image-wide statistics to calibrate predictions, as shown in Fig.~\ref{fig1}(a). Such coarse-grained strategies cannot diagnose whether errors originate from spurious correlations with non-causal visual attributes—such as texture, illumination, or sensor noise—which are often the underlying cause of failure under domain shifts. We argue that domain shifts can be effectively characterized as shifts in specific compositions of category-independent visual attributes, a perspective supported by language-based domain adaptation modeling~\cite{zeng2025explainingdomainshiftslanguage}. However, such adaptation techniques typically require offline attribute dictionary construction—often using large language models—followed by detector retraining. Counterfactual reasoning~\cite{sauer2021counterfactualgenerativenetworks, goyal2019counterfactualvisualexplanations, niu2021counterfactualvqacauseeffectlook, kusner2018counterfactualfairness, verma2022counterfactualexplanationsalgorithmicrecourses, goyal2019counterfactualvisualexplanations} offers a mechanism to explicitly identify and suppress misleading attribute dependencies by perturbing non-causal attributes in a controlled manner and constructing corresponding counterfactual samples. As shown in Fig.~\ref{fig1}(b), we propose directly observing how each region’s prediction responds to attribute-level changes. This allows us to quantify region-specific sensitivity to spurious visual cues and selectively suppress attribute-driven biases during inference. Thus, our approach extends causal adaptation principles into a practical, training-free TTA framework, enabling fine-grained and interpretable robustness to domain shifts.

Motivated by the above analysis, we propose a new and lightweight framework FACTOR, a counterFACtual TTA approach for Open-vocabulaRy object detection. Specifically, there are two steps: counterfactual probing and invariance-guided calibration.
In the first step, a perturbed counterpart of the test image is created by applying controlled transformations to predefined non-causal visual attributes (e.g., brightness, contrast, noise), preserving object semantics while altering appearance. Both original and counterfactual images are processed by the frozen detector, with their region proposals aligned spatially to form prediction pairs. In the next step, for each aligned region, we compute three signals: a counterfactual sensitivity score measuring prediction discrepancy, an attribute sensitivity score identifying specific attribute, and an attribute–category relevance score associates attribute with the corresponding category. These are fused into a region- and category-aware correction term, which is used to suppress spurious attribute-dependent predictions. In this way, FACTOR achieves fine-grained, attribute-aware adaptation without parameter updates, enabling interpretable and consistent performance gains across diverse corruption scenarios.

Our main contributions can be summarized as follows: (1) We propose FACTOR, a novel framework that improves robustness under domain shifts through explicit counterfactual reasoning. Unlike costly optimization or global calibration, FACTOR achieves efficient, region-aware TTA via attribute-level intervention and prediction invariance, filling a critical gap in fine-grained OVOD adaptation. (2) A counterfactual calibration mechanism is proposed that integrates region-level counterfactual sensitivity (CSS), attribute sensitivity (ASS) and attribute-category relevance (ACR) indicators. We use one multi-perturbed counterfactual sample, yet still disentangle region instability, attribute triggers, and category impact for targeted calibration.
(3) Extensive experiments on PASCAL-C, COCO-C, and FoggyCityscapes demonstrate that compared to existing TTA methods, FACTOR is state-of-the-art and generally achieves significant performance improvements.


\section{Related Work}

\textbf{Open-Vocabulary Object Detection (OVOD)} breaks the limitation of closed-set detection by enabling zero-shot generalization to novel categories, powered by vision-language pre-training that aligns visual and textual semantics. OVOD can be categorized into {\bf architectural alignment} and {\bf adaptation strategies}. Architecture-driven semantic alignment establishes the field by integrating pre-trained VLMs into detection pipelines through cross-modal grounding and dual-path Transformer architectures. For example, OWL-ViT~\cite{minderer2022simpleopenvocabularyobjectdetection} adapts CLIP with object proposal generation for open-vocabulary detection, and GLIP~\cite{li2022grounded} introduces a text-image grounding framework to enhance cross-modal alignment. Grounding DINO~\cite{devlin2019bert} further advances the field with a Transformer-based architecture integrating dual-path cross-attention, achieving state-of-the-art in-distribution performance. While these milestones achieve impressive zero-shot generalization, their performance is often anchored to the source distribution. To enhance robustness, {\bf adaptation strategies} have emerged. Representative works like CoOp~\cite{Zhou_2022} and CoCoOp~\cite{zhou2022conditionalpromptlearningvisionlanguage} introduce prompt learning to optimize textual templates via learnable tokens for better task-specific alignment. But they are limited to offline optimization and lack flexibility for dynamic domain shifts during inference.

\textbf{Test-time adaptation for OVOD.} TTA mitigates domain shift by dynamically adjusting model predictions to unlabeled OOD test samples, without retraining or target domain annotations. Existing TTA methods for OVOD fall into two main categories:
\textbf{Optimization-based TTA} aims to align OOD visual features with text embeddings through inference-time refinement, such as prompt tuning~\cite{belal2025vlod, gao2025testtimeadaptiveobjectdetection, zhang2024historical}, entropy minimization~\cite{belal2025vlod}, or mean-teacher frameworks with instance memory banks~\cite{gao2025testtimeadaptiveobjectdetection}. While effective in cross-domain scenarios, these methods inherently suffer from high computational latency and memory overhead due to the necessity of backpropagation and parameter maintenance during inference.
\textbf{Training-free TTA} prioritizes efficiency by employing lightweight calibration or feature fusion without parameter updates. Recent works~\cite{zhou2025bayesian,zhou2025bayesianarxiv,karmanov2024efficient} utilize Bayesian inference or cache-based mechanisms to dynamically update category priors and class embeddings. However, as these methods primarily operate on global statistics or confidence scores, they lack a principled mechanism to diagnose and suppress failures driven by spurious correlations with non-causal visual attributes—a gap that motivates our fine-grained, counterfactual intervention strategy.

\begin{figure*}[t]
\begin{center}
\centerline{\includegraphics[width=0.937\textwidth]{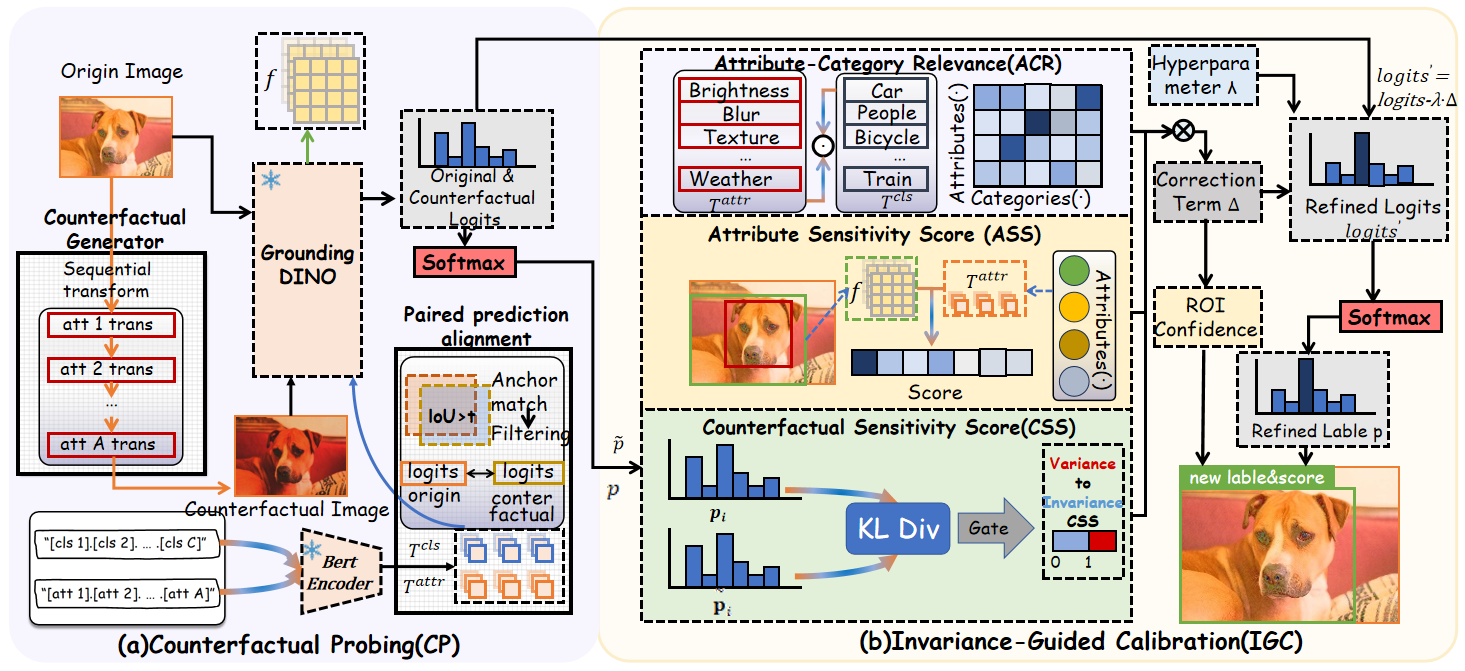}}
\caption{Overview of FACTOR. 
(a) \textbf{Counterfactual Probing (CP)}: A frozen Grounding DINO first processes the test image and its attribute-perturbed counterfactual image. The region predictions of the two views are then spatially aligned and paired with text-embedded attribute-category tokens.
(b) \textbf{Invariance-Guided Calibration (IGC)}: Counterfactual Sensitivity Score (CSS), Attribute Sensitivity Score (ASS), and Attribute-Category Relevance score (ACR) are fused to calculate category-specific correction terms to calibrate detection logits, thereby suppressing spurious feature dependencies and outputting refined detections.}
\label{icml-historical}
\end{center}
\vskip -0.3in
\end{figure*}

\section{Methodology}
\subsection{Problem Analysis}
{\bf Task definition.} Let $\mathcal{D}_S$ and $\mathcal{D}_T$ denote the source (pre-training) and target (test) data distributions, where $\mathcal{D}_T \neq \mathcal{D}_S$. Suppose that $\mathcal{C}$ is the comprehensive set of text embeddings representing all potential object categories. In the test-time adaptation setting, the model must adapt online using only a stream of unlabeled test samples $\{x_i\}_{i=1}^{\infty}$ from $\mathcal{D}_T$, without access to $\mathcal{D}_S$, source labels, or gradient-based optimization. The objective of this work is to maintain robust open-vocabulary detection performance of Grounding DINO~\cite{devlin2019bert} under distribution shift while preserving the model’s generalization to unseen categories.

{\bf Counterfactual invariance as a robustness criterion.} Distribution shifts in open-vocabulary detection primarily stem from spurious correlations learned during source-domain training. We conceptualize an image \(x\) as being generated from two latent factors: causal visual features \(z_c\) (e.g., object geometry, part composition) that are semantically essential for category recognition, and non-causal (spurious) features \(z_s\) (e.g., background context, global illumination, texture patterns, scale distribution) that are statistically associated with certain categories in the source domain \(\mathcal{D}_S\) but lack a causal link. A detector \(f_\theta\) (e.g., Grounding DINO) pre-trained on \(\mathcal{D}_S\) learns a mapping from image regions to category predictions. This often results in a decision function that inadvertently depends on both feature types. 
The dependency on \(z_s\) creates a vulnerability: when the correlation between \(z_s\) and category labels changes or disappears in the target domain \(\mathcal{D}_T\), detection performance degrades sharply. A robust open-vocabulary detector should base its predictions primarily on \(z_c\). This ideal can be formalized through the principle of counterfactual invariance: the detector’s output for a given region should remain stable under interventions that modify only non-causal image attributes while preserving semantic content. Let \(\text{do}(z_s = z_s')\) denote such an intervention, and let \(x' = \mathcal{T}(x)\) be the resulting counterfactual image obtained via a transformation \(\mathcal{T}\) that primarily affects \(z_s\). An invariant detector \(f^*\) satisfies:
\[
P_{f^*}(y \mid x) = P_{f^*}(y \mid x') \quad \forall y,
\]
or equivalently,
\[
\mathcal{D}_\text{KL}\left( P_{f^*}(y \mid x) \parallel P_{f^*}(y \mid x') \right) = 0.
\]
In practice, the frozen pre-trained detector \(f_\theta\) violates this invariance because it relies on \(z_s\). Our core insight is that the magnitude of this violation—measurable as the prediction discrepancy between \(x\) and \(x'\)—directly indicates the extent of spurious feature reliance for each region proposal. Consequently, test-time adaptation can be reformulated as promoting approximate counterfactual invariance through output calibration, without gradient updates or source data. By perturbing \(x\) in ways that predominantly affect suspected \(z_s\) (e.g., via noise, gamma, scaling), we obtain an invariance signal that guides the suppression of unreliable, shift-sensitive predictions.

\textbf{Overview of our method.} Based on the above analysis, the proposed method FACTOR is illustrated in Fig~\ref{icml-historical}, which consists of two steps. First, a counterfactual version of the test image through controlled attribute perturbations is generated in the step of counterfactual probing (Sec. 3.2). Then, in the step of invariance-guided calibration (Sec.3.3), by comparing the region-level predictions between the original and its counterfactual counterpart, we quantify prediction discrepancy and region-wise attribute sensitivity and semantic relevance to compute calibration signals. These signals are then fused into a correction term that directly adjusts the detector's category logits. This suppresses attribute-driven false predictions while preserving reliable semantic cues. The entire process requires no backpropagation or parameter updates, relied only on a single counterfactual forward pass, thereby achieving efficient, interpretable, and region-aware adaptation. 

\subsection{Counterfactual Probing}
This step aims to expose the model’s dependence on spurious features by contrasting its predictions under controlled attribute perturbations. We probe its decision behavior by comparing region-level predictions between the original image and a minimally perturbed counterfactual counterpart. This allows us to directly identify prediction variations triggered by non-causal attributes while preserving efficiency and requiring no parameter updates. It consists of two key steps: counterfactual image construction and paired prediction alignment, which together provide a lightweight yet effective mechanism to assess and address instability during inference.

\subsubsection{Counterfactual image construction}
Given a test image $x$, our goal is to construct a counterfactual image $x^{\mathrm{cf}}$ that preserves semantic content while selectively perturbing non-causal visual attributes. We intervene on a predefined set of interpretable attributes $\mathcal{A}$ that commonly induce spurious correlations in open-vocabulary object detection, including brightness, contrast, blur, noise, texture, and weather conditions. Each attribute $a\in\mathcal{A}$ is associated with a deterministic transformation operator $\mathcal{T}_a(\cdot)$.
Specifically, we define $\mathcal{T}_{\text{brit}}(x)$, $\mathcal{T}_{\text{cont}}(x)$, $\mathcal{T}_{\text{blur}}(x)$, $\mathcal{T}_{\text{noise}}(x)$, $\mathcal{T}_{\text{text}}(x)$ and $\mathcal{T}_{\text{weath}}(x)$ as corresponding to different counterfactual operations; the specific formulas can be found in the Appendix.
This design guarantees that any prediction discrepancy between $x$ and $x^{\mathrm{cf}}$ can be attributed to the model’s sensitivity to attribute-level shifts rather than to content corruption.
We compose all attribute transformations sequentially to generate a single counterfactual image:
\begin{equation}
x^{\mathrm{cf}} = \mathcal{T}(x) = \bigcirc_{a \in \mathcal{A}} \mathcal{T}_a(x),
\label{eq:cf_compose}
\end{equation}
where $\bigcirc$ denotes sequential application. Applying all attributes to the same image enables a comprehensive probe of multi-factor distribution shifts while requiring only a single additional forward pass through the frozen detector, leading to a significant reduction in inference time. By preserving spatial structure and object identity and perturbing only attribute-level appearance, any prediction discrepancy between $x$ and $x^{\mathrm{cf}}$ can be directly attributed to sensitivity to non-causal visual attributes, providing a principled basis for counterfactual probing and test-time calibration.

\subsubsection{Paired Prediction Alignment}
For subsequent invariance comparison and logit-level calibration, we need to construct spatially and semantically grounded prediction pairs. We forward both the original image $x$ and the counterfactual image $x^{\mathrm{cf}}$ through the frozen open-vocabulary detector, obtaining region proposals $\mathcal{R}=\{r_i\}_{i=1}^{N}$ and $\tilde{\mathcal{R}}=\{\tilde{r}_j\}_{j=1}^{\tilde{N}}$, respectively.
To enforce spatial invariance under attribute perturbations, each region $r_i\in\mathcal{R}$ is matched to the counterfactual region with the highest Intersection-over-Union (IoU), and matches with IoU below $0.3$ are discarded.
For each aligned region pair $(r_i,\tilde{r}_i)$, we get the corresponding category-level paired predictions that differ only in attribute-level appearance. This alignment removes geometric variability and isolates prediction changes caused by non-causal attribute shifts.

To support semantically meaningful calibration under distribution shifts, we embed both attribute tokens and category names into a shared semantic space using Grounding DINO’s BERT encoder~\cite{devlin2019bert}, producing attribute embeddings $\mathbf{T}^{\mathrm{attr}}\in\mathbb{R}^{|\mathcal{A}|\times D}$ and category embeddings $\mathbf{T}^{\mathrm{cls}}\in\mathbb{R}^{|C|\times D}$.
Together, spatially aligned prediction pairs and shared semantic representations provide a clean and interpretable foundation for subsequent invariance estimation and attribute-aware logit calibration.

\subsection{Invariance-Guided Calibration}
Building on the paired predictions obtained through counterfactual probing, we introduce an invariance-guided calibration mechanism that adjusts the detector’s outputs by quantifying their sensitivity to attribute perturbations. This stage comprises three complementary metrics, which are then fused into a correction term and applied via logit adjustment to achieve robust prediction calibration.

\subsubsection{Invariance-Aware Sensitivity Metrics}
To diagnose a detector’s sensitivity to spurious correlations—without ground-truth supervision—we introduce a tri-level metric framework that progressively disentangles prediction discrepancy. First, the Counterfactual Sensitivity Score (CSS) quantifies whether a region’s prediction is invariant under counterfactual perturbations. Second, the Attribute Sensitivity Score (ASS) identifies which visual attribute contributes to that discrepancy by measuring region–attribute alignment. Third, the Attribute–Category Relevance (ACR) provides a semantic prior on how strongly each attribute influences each category. 

{\bf Counterfactual Sensitivity Score (CSS).} For each aligned region pair $(r_i, \tilde{r}_i)$, we get the category probability distributions $\mathbf{p}_i \in \mathbb{R}^C$ (from the original image $x$) and $\tilde{\mathbf{p}}_i \in \mathbb{R}^C$ (from the counterfactual image $x^{\mathrm{cf}}$) from the detector outputs. To quantify the raw prediction discrepancy induced by attribute perturbations, we compute the Kullback–Leibler (KL) divergence~\cite{shlens2014noteskullbackleiblerdivergencelikelihood} between these distributions:
\begin{equation}
\mathrm{KL}_i = \sum_{c=1}^{C} p_i(c)\log\frac{p_i(c)}{\tilde{p}_i(c)},
\label{eq:raw_kl}
\end{equation}
where $\mathrm{KL}_i \geq 0$ with equality if and only if the predictions are invariant to counterfactual perturbations. By using the forward form (weighted by $\mathbf{p}_i$), we prioritize discrepancy in categories where the detector initially had high confidence—ensuring we focus on meaningful prediction shifts rather than low-probability noise. A non-zero $\mathrm{KL}_i$ indicates sensitivity to attribute perturbations, while larger values reflect increasing prediction discrepancy. The KL divergence is computed once per region and shared across all attributes, making it computationally efficient.

However, raw $\mathrm{KL}_i$ values are overly sensitive to trivial prediction variations. To suppress calibration for minor fluctuations while retaining it for pronounced discrepancy, we gate the KL divergence with an adaptive, smooth threshold. Let
$\mu = \frac{1}{N}\sum_{j=1}^{N} \mathrm{KL}_j$ be the average KL, which serves as a baseline for ``normal'' variation. Here, $N$ represents the total number of paired regions in the test image. We then define Counterfactual Sensitivity Score (CSS) as
\begin{equation}
\mathrm{CSS}_i = \sigma\!\bigl(\mathrm{KL}_i - \mu\bigr),
\end{equation}
where $\sigma$ is the sigmoid function. This design yields a soft 0--1 weight: $\mathrm{CSS}_i \approx 1$ when $\mathrm{KL}_i$ is well above the batch average (severe discrepancy), and $\mathrm{CSS}_i \approx 0$ when it is close to or below the average (benign variation). Consequently, calibration is applied predominantly to regions that exhibit substantial sensitivity to the counterfactual perturbation.

{\bf Attribute Sensitivity Score (ASS).} While CSS quantifies perdition discrepancy in a region, it does not reveal which attributes are likely responsible. To model region-wise sensitivity to individual attributes, we introduce ASS. 
Given the visual feature $\mathbf{f}_i \in \mathbb{R}^D$ of region $r_i$ and the $a$-th attribute embedding $\mathbf{T}^{\mathrm{attr}}_a \in \mathbb{R}^D$, ASS is defined as:
\begin{equation}
\mathrm{ASS}_{i,a}
=
\sigma\!\left(\mathbf{f}_i^{\top}\mathbf{T}^{\mathrm{attr}}_a\right),
\label{eq:asf}
\end{equation}
where $\sigma(\cdot)$ denotes the sigmoid function, which maps its input to the interval $[0, 1]$. ASS measures the alignment between region appearance and an attribute’s semantic direction in the shared vision-language space supported by Grounding DINO text encoder.
A higher $\mathrm{ASS}_{i,a}$ indicates that region $r_i$ exhibits visual characteristics strongly associated with attribute $a$, suggesting that its predictions are more likely to be affected by perturbations along that attribute dimension.

{\bf Attribute-Category Relevance (ACR) score.} Even when a region is sensitive to a particular attribute, not all categories are equally affected by that attribute. To quantify category-dependent relevance, we define ACR as a purely semantic measure based on text embeddings as
\begin{equation}
\mathrm{ACR}_{a,c}
=
\sigma\!\left((\mathbf{T}^{\mathrm{attr}}_a)^{\top}\mathbf{T}^{\mathrm{cls}}_c\right),
\label{eq:acr}
\end{equation}
where $\mathbf{T}^{\mathrm{cls}}_c \in \mathbb{R}^D$ denotes the text embedding of category $c$.
ACR quantifies the semantic affinity between attribute $a$ and category $c$ in the language embedding space.
A high $\mathrm{ACR}_{a,c}$ implies that attribute $a$ is conceptually relevant to category $c$, providing a principled prior for category-aware calibration.
Notably, ACR is independent of image content and is shared across all regions, reflecting stable linguistic structure rather than instance-level appearance.

\subsubsection{Logit adjustment and inference}

\textbf{Correction term.} Three metrics capture distinct yet complementary aspects of distribution shift. Their combination enables fine-grained and invariance-aware calibration, while keeping metric computation disentangled from correction and optimization. Combining $\mathrm{CSS}_i$, $\mathrm{ASS}_{i,a}$, and $\mathrm{ACR}_{a,c}$, the category-wise correction for $r_i$ and $c$ is:
\begin{equation}
\Delta_i(c) = \frac{1}{|\mathcal{A}|} \sum_{a \in \mathcal{A}} \mathrm{ASS}_{i,a} \cdot \mathrm{ACR}_{a,c} \cdot \mathrm{CSS}_i.
\label{eq:attribute_sensitivity}
\end{equation}
Averaging over $\mathcal{A}$ balances correction across shift factors. 
These three components form a joint diagnostic chain for precise adaptation: $\mathrm{CSS}_i$ acts as a gate to detect whether the prediction in a region is inherently unstable, while the synergy between $\mathrm{ASS}_{i,a}$ and $\mathrm{ACR}_{a,c}$ provides the necessary directionality by determining which attribute caused the error and which category should be suppressed. This structured intervention allows FACTOR to mitigate spurious attribute-dependent predictions while preserving robust, semantically-grounded cues.

{\bf Detection calibration.} We calibrate the detector's logits for $r_i$ as:
\begin{equation}
\mathbf{logits}_i' = \mathbf{logits}_i - \lambda \cdot \Delta_i,
\label{eq:attribute_sensitivity}
\end{equation}
where $\lambda > 0$ is a hyperparameter that controls adaptation strength. The subtraction penalizes categories unstable under counterfactual perturbations. Updated probabilities are $\mathbf{p}_i' = \sigma(\mathbf{logits}_i')$. To further account for region robustness, we compute an ROI-level confidence score:
$\bar{\Delta}_i = \sum_{c} p_i'(c)\, \Delta_i(c)$, and refine the detection confidence as:
\begin{equation}
s_i'= s_i \cdot \exp\!\left(-\bar{\Delta}_i\right),
\end{equation}
this step down-weights counterfactually discrepancy regions while preserving consistent detections.

\begin{table*}[t]
\centering
\caption{Object detection performance comparison with the state-of-the-art methods on FoggyCityscapes.
Metric: Average Precision@50(\%) - the best numbers are denoted in {\bf bold};
Bp-free: backpropagation-free at test time.}
\label{tab:FoggyCityscapes}
\resizebox{\textwidth}{!}{
\begin{tabular}{l|ccc|c c c c c c c c|c}
\hline
Methods &Venue &Framework &Bp-free&Pson & Rder & Car & Tuck & Bus & Train & Mcle & Bcle & $\text{mAP}_{50}$  \\
\hline
\multicolumn{13}{c}{\textbf{Visual Backbone: ResNet-50 }}  \\ \hline
SHOT \cite{liang2020we} &ICML20&Faster RCNN&  $\XSolidBrush$& 26.7 & 30.3 & 36.9 & \bf16.8 & \underline{28.9} & \underline{6.4} & 14.3 & 23.3 & 23.0 \\
T3A \cite{iwasawa2021test} &NIPS21&Faster RCNN& $\XSolidBrush$& 22.6 & 23.0 & 31.9 & 7.7 & 14.8 & 1.0 & 7.9 & 19.7 & 16.6 \\
Self-Training \cite{xu2021end} &CVPR21&Faster RCNN& $\XSolidBrush$& \underline{27.7} & \underline{30.8} & \underline{41.4} & 12.8 & 27.4 & 4.2 & \underline{14.8} & \underline{26.1} & \underline{23.1} \\
TTAC \cite{su2022revisiting} &NIPS22&Faster RCNN& $\XSolidBrush$& 24.5 & 27.3 & 33.4 & 14.6 & 26.1 & 5.8 & 14.1 & 21.5 & 20.9 \\
STFAR\cite{chen2023stfar}  &ARXIV23&Faster RCNN& $\XSolidBrush$& \bf28.8 & \bf32.0 & \bf42.4 & \underline{15.1} & \bf30.1 & \bf11.2 & \bf15.5 & \bf26.2 & \bf25.1 \\
\hline
\multicolumn{13}{c}{\textbf{Visual Backbone: Swin-T}}  \\ \hline
GDINO \cite{liu2024grounding}&ECCV24&Grounding DINO& $\CheckmarkBold$ &30.10&3.42&46.26&22.41&31.98&0.08&27.96&28.87&23.88 \\
TDA \cite{karmanov2024efficient} &CVPR24&Grounding DINO& $\CheckmarkBold$&35.27&4.87&46.43&22.82&32.30&0.38&28.68&29.86&25.08\\
HisTPT \cite{zhang2024historical} &NIPS24 &Grounding DINO & $\XSolidBrush$&32.50&5.09&47.45&\underline{23.13}&34.63&0.85&28.82&31.98&25.55\\
BCA \cite{zhou2025bayesian} &CVPR25&Grounding DINO& $\CheckmarkBold$ & \underline{41.01}&\underline{5.61}&\underline{47.85}&21.86&33.40&\bf0.91&28.47&29.82&26.12\\
BCA+ \cite{zhou2025bayesianarxiv}&ARXIV25&Grounding DINO& $\CheckmarkBold$&39.03&3.61&46.98&22.26&\underline{35.26}&\underline{0.87}&\underline{29.37}&\underline{35.87}&\underline{26.65}  \\
\hline
FACTOR & Ours & Grounding DINO & $\checkmark$ &  \textbf{43.39} & \textbf{8.18} & \textbf{56.38} & \textbf{26.98} & \textbf{39.02} & 0.52 & \textbf{30.70} & \textbf{46.36} & \textbf{31.44} \\
\hline
\multicolumn{13}{c}{\textbf{Visual Backbone: Swin-B}}  \\ \hline
GDINO \cite{liu2024grounding} &ECCV24&Grounding DINO& $\CheckmarkBold$&34.95&20.35&51.56&30.01&45.72&0.87&35.21&32.06&31.34  \\
{TDA} \cite{karmanov2024efficient} &CVPR24&Grounding DINO& $\CheckmarkBold$ 
&\underline{38.29}&\underline{25.53}&50.09&30.49&\underline{45.92}&11.98&33.12&39.06&34.31
\\
HisTPT \cite{zhang2024historical} &NIPS24 &Grounding DINO & $\XSolidBrush$ &37.95&22.46&\underline{54.00}&29.98&44.11&12.50&\underline{37.27}&42.33&35.07\\
BCA \cite{zhou2025bayesian} &CVPR25&GroundingDINO& $\CheckmarkBold$&36.53&24.42&50.77&\underline{30.83}&45.25&\textbf{20.29}&33.20&\underline{42.49}&35.47
\\
BCA+ \cite{zhou2025bayesianarxiv}&ARXIV25&Grounding DINO& $\CheckmarkBold$
&38.15&\textbf{26.30}&52.30&30.77&45.66&\textbf{20.29}&33.89&42.41&\underline{36.22}\\

\hline
FACTOR  & Ours & Grounding DINO & $\checkmark$ & \textbf{42.55} & 22.36 & \textbf{60.64} & \textbf{32.97} & \textbf{49.03} & \underline{17.05} & \textbf{38.19} & \textbf{50.67} & \textbf{39.18} \\
\hline
\end{tabular}}
\end{table*}

\begin{table*}[t]
\centering
\caption{Object detection performance comparison with the state-of-the-art methods on PASCAL-C.
Metric: mean Average Precision@50(\%) -  the best numbers are denoted in {\bf bold}.}
\label{tab:pascalc}
\resizebox{\textwidth}{!}{
\begin{tabular}{l|c c c c c c c c c c c c c c c|c}
\hline
{Methods} & Brit & Contr & Defoc & Elast & Fog & Frost & Gauss & Glass & Impul & Jpeg & Motn & Pixel & Shot & Snow & Zoom & Average \\
\hline
\multicolumn{17}{c}{\textbf{Visual Backbone: ResNet-50 }}  \\ \hline
SHOT \cite{liang2020we} & \underline{72.0} & 31.7 & 18.9 & 46.6 & 67.5 & 45.8 & 12.0 & 11.6 & 16.4 & 41.8 & 19.7 & 33.1 & 19.9 & 42.5 & 27.6 & 33.8 \\
T3A \cite{iwasawa2021test} & 36.9 & 12.5 & 11.0 & 19.7 & 32.7 & 20.6 & 6.1 & 6.4 & 6.5 & 14.8 & 10.1 & 13.2 & 8.4 & 16.8 & 13.8 & 15.3 \\
Self-Training \cite{xu2021end} & 67.9 & 39.3 & 2.6 & 52.5 &\underline{65.7} & 47.2 & 11.9 & 20.2 & 12.1 & 29.3 & 4.1 & 6.9 & 17.4 & 44.9 & 9.5 & 28.8 \\
TTAC \cite{su2022revisiting} & \bf72.2 & \underline{40.4} & \underline{29.3} & \bf58.1 & \bf68.7 & \underline{50.4} & \underline{29.8} & \underline{28.7} & \underline{33.6} & \underline{46.4} & \underline{29.2} & \underline{46.1} & \underline{35.1} & \underline{48.0} & \bf34.9 & \underline{43.4} \\
STFAR \cite{chen2023stfar} & 67.3 & \bf51.8 & \bf34.8 & \underline{55.7} & 65.2 & \bf50.7 & \bf32.4 & \bf34.6 & \bf36.3 & \bf49.4 & \bf34.6 & \bf55.7 & \bf37.8 & \bf50.9 & \underline{34.8} & \bf46.1 \\
\hline
\multicolumn{17}{c}{\textbf{Visual Backbone: Swin-T}}  \\ \hline
GDINO \cite{liu2024grounding} &63.04&45.66&35.58&31.33&60.92&50.52&25.82&18.84&27.99&39.34&26.99&13.39&30.49&45.89&23.38&35.95  \\
TDA \cite{karmanov2024efficient} &64.61&47.71&37.33&32.41&61.77&52.19&27.05&20.48&29.35&41.45&28.50&14.15&31.85&48.07&24.35&37.42\\
HisTPT \cite{zhang2024historical} &66.27&48.10&38.71&34.32&64.24&53.46&\underline{28.68}&21.75&30.68&41.56&30.20&\underline{15.90}&33.47&48.46&25.52&38.76\\
{BCA} \cite{zhou2025bayesian} &68.11&49.42&38.46&34.42&64.27&53.21&28.37&22.24&\underline{31.90}&42.81&30.39&14.81&\underline{33.68}&49.37&24.97&39.10 \\
BCA+ \cite{zhou2025bayesianarxiv}
&\underline{70.25}&\underline{50.72}&\underline{40.79}&\underline{37.08}&\underline{67.83}&\underline{56.38}&28.59&\underline{22.48}&31.74&\underline{47.40}&\underline{30.75}&14.87&33.66&\underline{51.87}&\underline{25.65}&\underline{40.67} \\ 

 \hline
FACTOR (Ours) & \textbf{74.46} & \textbf{55.42} & \textbf{45.33} & \textbf{41.78} & \textbf{72.63} & \textbf{61.39} & \textbf{33.97} & \textbf{26.54} & \textbf{37.02} & \textbf{49.45} & \textbf{36.09} & \textbf{19.42} & \textbf{38.99} & \textbf{56.77} & \textbf{28.60} & \textbf{45.19} \\
\hline
\multicolumn{17}{c}{\textbf{Visual Backbone: Swin-B}}  \\ \hline
GDINO  \cite{liu2024grounding}
&85.16&70.69&56.12&55.56&84.56&73.75&54.94&41.17&57.72&71.44&56.08&66.17&60.54&76.48&36.94&63.15
\\
TDA \cite{karmanov2024efficient} &86.28&74.33&57.98&58.45&85.98&76.41&57.82&43.84&60.80&74.31&57.91&69.45&63.79&78.36&38.84&65.64
\\
HisTPT \cite{zhang2024historical} & 89.33&74.89&60.21&59.53&88.14&78.22&58.32&45.48&61.00&75.93&60.13&69.69&64.69&79.79&40.80&67.07\\
{BCA} \cite{zhou2025bayesian} &88.66&75.73&59.84&61.00&88.02&77.69&60.21&45.98&62.46&76.32&60.55&71.22&64.89&80.67&40.66   &67.59
\\
BCA+ \cite{zhou2025bayesianarxiv}
&\underline{89.39}&\underline{77.85}&\underline{62.04}&\underline{62.71}&\underline{88.87}&\underline{79.94}&\underline{61.81}&\underline{47.67}&\underline{64.87}&\underline{78.12}&\underline{61.81}&\underline{73.51}&\underline{67.28}&\underline{82.42}&\underline{41.42}&\underline{69.31}
\\

\hline
FACTOR (Ours) & \textbf{92.86} & \textbf{84.20} & \textbf{69.35} & \textbf{70.70} & \textbf{92.80} & \textbf{85.68} & \textbf{71.40} & \textbf{57.87} & \textbf{74.28} & \textbf{83.22} & \textbf{69.71} & \textbf{81.23} & \textbf{75.68} & \textbf{87.45} & \textbf{48.11} & \textbf{76.30} \\
\hline
\end{tabular}}
\end{table*}

\section{Experiments}
\subsection{Experimental Setup}
\textbf{Datasets.}
Consistent with existing TTA methods~\cite{zhou2025bayesian}, we evaluate FACTOR on three established OVOD test-time adaptation benchmarks that simulate real-world distribution shifts through image corruption. FoggyCityscapes \cite{sakaridis2018semantic} is a synthetic foggy variant of Cityscapes \cite{cordts2016cityscapes}, evaluated at the most challenging fog density ($\beta$=0.02) causing severe visibility degradation. PASCAL-C \cite{michaelis2019benchmarking} applies 15 common corruptions to the PASCAL VOC 2007 test set \cite{everingham2015pascal}, evaluated at severity level 5. COCO-C \cite{michaelis2019benchmarking} introduces the same 15 corruptions to 5,000 images from MS-COCO 2017 validation set \cite{lin2014microsoft}. The 15 corruptions are: brightness, contrast, defocus, elastic, fog, frost, gaussian, glass, impulse, jpeg, motion, pixelate, shot, snow, zoom. The following text uses abbreviations, such as brightness abbreviated as Brit. Together, these benchmarks cover adverse weather (FoggyCityscapes) and diverse synthetic corruptions (PASCAL-C, COCO-C), providing a comprehensive testbed for robustness evaluation.

\textbf{Implementation details.}
All experiments are conducted with PyTorch (batch size = 1) under the TTA protocol, using Grounding DINO as the base detector (Swin-T and Swin-B\cite{liu2021swin} visual backbones, BERT text encoder). Following BCA+ \cite{zhou2025bayesianarxiv}, we adopt the AP50 as our primary metric. Category names are encoded via the prompt template ``\textit{(class 1). (class 2). $\cdots$ (class $C$).}''\cite{zhou2025bayesianarxiv}, and six interpretable visual attributes (brightness, contrast, blur, noise, texture, weather) are encoded with the same BERT encoder and prompt format. Attribute/category names are embedded into a shared semantic space using the pretrained text encoder, with no source data/annotations/supervision used at test time. For each test image, we run FACTOR one forward pass on the original image and one on the composed counterfactual image to estimate attribute-induced prediction divergence efficiently. The hyperparameters are fixed across all datasets: $\lambda = 0.5$, and a fixed attribute set for counterfactual composition.

\begin{table*}[t]
\centering
\caption{Object detection performance comparison with the state-of-the-art methods on COCO-C.
Metric: mean Average Precision@50(\%) -  the best numbers are denoted in {\bf bold}.}
\label{tab:cococ}
\resizebox{\textwidth}{!}{
\begin{tabular}{l|c c c c c c c c c c c c c c c|c}
\hline
{Methods} & Brit & Contr & Defoc & Elast & Fog & Frost & Gauss & Glass & Impul & Jpeg & Motn & Pixel & Shot & Snow & Zoom & Average \\\hline
\multicolumn{17}{c}{\textbf{Visual Backbone: ResNet-50}}  \\ \hline
SHOT \cite{liang2020we} & \bf40.9 & 26.6 & 14.7 & 19.7 & \bf41.5 & 26.7 & 11.0 & 7.2 & 12.1 & 16.4 & 11.0 & 9.7 & 13.0 & 22.0 & 6.4 & 18.6 \\
T3A \cite{iwasawa2021test} & 28.8 & 15.9 & 8.3 & 11.3 & 28.9 & 17.2 & 4.6 & 3.1 & 5.2 & 9.0 & 5.8 & 4.1 & 5.8 & 13.8 & 3.5 & 11.0 \\
Self-Training \cite{xu2021end}& 38.1 & 28.4 & 14.7 & 25.5 & 38.5 & 27.9 & 16.7 & 11.4 & \underline{18.8} & \underline{23.8} & \underline{16.0} & 24.5 & 18.6 & \underline{27.6} & 7.8 & 22.6 \\
TTAC \cite{su2022revisiting} & 38.3 & \underline{29.5}& \underline{15.1} & \underline{28.2} & \underline{39.0} & \underline{28.5} & \underline{16.8} & \underline{14.3} & 18.0 & 23.2 & 14.3 & \underline{24.8} & \underline{19.3} & 26.7 & \underline{8.7} & \underline{23.0} \\
STFAR \cite{chen2023stfar} & \underline{39.1} & \bf31.1 & \bf16.8 & \bf29.0 & \underline{39.0} & \bf29.2 & \bf19.2 & \bf15.4 & \bf20.1 & \bf26.1 & \bf17.2 & \bf28.3 & \bf21.0 & \bf29.5 & \bf10.2 & \bf24.7 \\
W3TTAOD \cite{yoo2024and} & 36.4 & 27.2 & 14.0 & 27.2 & 37.4 & 27.2 & 13.6 & 13.6 & 16.1 & 22.3 & 14.2 & 22.2 & 16.6 & 23.7 & 8.3 & 21.3 \\
\hline

\multicolumn{17}{c}{\textbf{Visual Backbone: Swin-T}}  \\ \hline
GDINO \cite{liu2024grounding} &42.91&23.83&19.45&23.76&43.26&30.36&15.03&10.22&15.87&23.77&16.56&7.31&16.99&25.51&9.35& 21.61 \\
TDA \cite{karmanov2024efficient} &45.53 & 24.43 & 20.33 & 27.08 & 45.91 & 32.31 & 15.77 & 9.93 & 16.72 & 26.25 & 16.85 & \underline{8.99} & 17.10 & 28.22 & 9.62&23.00\\
HisTPT \cite{zhang2024historical} &45.55&26.07&21.25&26.02&45.09&32.82&17.55&\underline{12.45}&18.21&26.16&\underline{19.12}&8.62&17.88&28.19&9.93&23.66\\
{BCA} \cite{zhou2025bayesian} &46.34 & 25.18 & 21.03 & 27.45 & 46.74 & 33.92 & 15.97 & 10.91 & 17.33 & 26.19 & 16.05 & 8.89 & 17.97 & 29.02 & 9.87& 23.52  \\
BCA+ \cite{zhou2025bayesianarxiv}
&\underline{49.62}&\underline{26.56}&\underline{22.20}&\underline{28.65}&\underline{49.79}&\underline{35.65}&\underline{17.97}&11.47&\underline{18.67}&\underline{28.38}&19.10&7.83&\underline{19.86}&\underline{30.10}&\underline{10.08}&\underline{25.06}\\

\hline
FACTOR (Ours) & \textbf{54.71} & \textbf{32.15} & \textbf{25.99} & \textbf{32.69} & \textbf{55.09} & \textbf{41.04} & \textbf{21.76} & \textbf{14.76} &\textbf{ 23.33} & \textbf{31.84} & \textbf{22.22} & \textbf{12.87} & \textbf{23.98} & \textbf{34.98} &\textbf{12.05} & \textbf{29.30} \\
\hline
\multicolumn{17}{c}{\textbf{Visual Backbone: Swin-B}}  \\ \hline

GDINO \cite{liu2024grounding} 
&55.36&40.40&29.11&36.51&56.26&44.39&30.27&21.38&31.30&40.26&28.78&36.68&32.67&42.71&13.44&35.97
\\
TDA \cite{karmanov2024efficient} 
&56.86&42.26&30.35&38.15&58.21&46.72&32.28&22.64&32.84&42.12&29.94&38.97&34.17&44.14&14.89&37.64
\\
HisTPT \cite{zhang2024historical} &58.02&43.28&31.37&38.51&59.09&46.96&32.85&23.53&33.92&42.23&31.28&39.45&35.06&44.72&\underline{15.57}&38.39\\
{BCA} \cite{zhou2025bayesian} 
&57.21&42.32&31.00&38.90&58.84&46.43&32.17&23.37&33.53&42.49&31.15&39.56&35.76&45.42&14.92&38.20\\
BCA+ \cite{zhou2025bayesianarxiv}
&\underline{60.34}&\underline{44.99}&\underline{32.25}&\underline{40.85}&\underline{60.92}&\underline{49.06}&\underline{34.04}&\underline{24.42}&\underline{35.33}&\underline{44.91}&\underline{31.72}&\underline{41.17}&\underline{36.59}&\underline{47.85}&15.24&\underline{39.98}
\\
\hline
FACTOR (Ours) & \textbf{65.89} & \textbf{51.57} & \textbf{37.17} & \textbf{46.07} & \textbf{66.98} & \textbf{55.77} & \textbf{40.57} & \textbf{30.31} & \textbf{42.13} & \textbf{49.79} & \textbf{36.86} & \textbf{48.06} & \textbf{43.16} & \textbf{53.72} & \textbf{18.66} & \textbf{45.78} \\
\hline
\end{tabular}}
\vspace{-0.4cm}
\end{table*}

\subsection{Comparisons with State-of-the-Arts}
We compare FACTOR with state-of-the-art test-time adaptation methods. Traditional approaches built upon closed-vocabulary detectors, such as Faster R-CNN with ResNet-50 backbones, include SHOT~\cite{liang2020we}, T3A~\cite{iwasawa2021test}, Self-Training~\cite{xu2021end}, TTAC~\cite{su2022revisiting}, STFAR~\cite{chen2023stfar}, and W3TTAOD~\cite{yoo2024and} (on COCO-C only). More recent methods leverage VLMs, particularly Grounding DINO, including GDINO~\cite{liu2024grounding}, TDA~\cite{karmanov2024efficient}, HisTPT~\cite{zhang2024historical}, BCA~\cite{zhou2025bayesian}, and BCA+~\cite{zhou2025bayesianarxiv}.

\textbf{Quantitative  comparison.} Tables \ref{tab:FoggyCityscapes}-\ref{tab:Efficiency} demonstrate that FACTOR consistently achieves state-of-the-art performance across all three benchmarks—FoggyCityscapes, PASCAL-C, and COCO-C—under both Swin-T and Swin-B backbones. We have the following observations:
(1) {\bf Broad improvements with particular gains on visually sensitive objects.} Across datasets with fundamentally different corruption characteristics, FACTOR delivers stable and substantial improvements over prior methods, indicating strong robustness to diverse and severe domain shifts.
Instead of concentrating gains on a small subset of easy classes, FACTOR also exhibits broad improvements, particularly for objects whose appearance is highly susceptible to visibility degradation, texture distortion, or noise. For example, while Grounding DINO suffers significant drops in "Pixel" and "Glass", FACTOR consistently recovers performance across these varied categories, outperforming the baseline by 6.03\% and 7.7\% AP50 on PASCAL-C (SwinT, Table \ref{tab:pascalc}), respectively. This suggests that the proposed counterfactual calibration effectively suppresses spurious attribute cues that are amplified under domain shifts, rather than merely boosting confidence for already reliable predictions. 

(2) {\bf Satisfactory results on hard categories.} FACTOR shows suboptimal results for "Train" and "Rider" in Table \ref{tab:FoggyCityscapes}. We attribute this to FACTOR's instance-based nature: it prioritizes local counterfactual invariance per image, whereas BCA+ benefits from accumulating historical statistics and spatial priors across the test sequence. Nevertheless, FACTOR still achieves satisfactory performance on these challenging classes. Enhancing region-aware calibration with target-domain knowledge accumulation remains a promising research direction.

(3) {\bf Balance between computational cost and performance.} As shown in Table~\ref{tab:Efficiency}, FACTOR achieves a favorable balance between performance gains and computational cost at test time. Compared with existing methods, FACTOR introduces only a modest inference overhead while delivering consistently stronger improvements under domain shifts. Notably, methods with higher computational complexity do not exhibit proportional performance benefits, whereas FACTOR attains superior robustness with substantially lower cost. These results indicate that the proposed calibration strategy is not only effective but also computationally efficient for practical test-time deployment. 
Overall, the above mentioned findings confirm that counterfactual, attribute-guided calibration constitutes a principled and robust solution for test-time adaptation in open-vocabulary object detection.
\begin{table}[t]
\centering
\caption{Efficiency and performance comparison on COCO-C Brit. Backbone: Swin-T. GPU: NVIDIA A100. Time: average inference time (ms) for 5,000 images. Mem: GPU memory (MB).}
\label{tab:Efficiency}
\resizebox{0.48\textwidth}{!}{
\small
\begin{tabular}{l|c c c}
\hline
{Method} & $\text{mAP}_{50}$ & Time(ms)  & Mem(MB) \\ \hline
GDINO\cite{liu2024grounding}  &  42.91 &147.6& 3025\\
TDA\cite{karmanov2024efficient}  &45.53& 411.6 &4586\\
HisTPT\cite{zhang2024historical}  &45.55& 1240.8 &12850\\
BCA+\cite{zhou2025bayesianarxiv}  &49.62 &190.8 &3862 \\ 

\hline
FACTOR & 54.71 & 286.3 & 3097 \\
\hline
\end{tabular}}
\vspace{-0.2cm}
\end{table}

\textbf{Qualitative  comparison.} We selected two representative images to demonstrate the visual comparison results. The second best method BCA+ is compared. These images are from the COCO-C Brit(Brightness) dataset, and the backbone is SwinT. The first image denotes a scene with dense objects, including small objects with blurred features. FACTOR successfully identified most of the objects; while the second image represents a scene with strong lighting interference. The sunlight severely damaged the features of the target person and skis. Although BCA+ is able to identify the skis compared to GroundingDINO, it could not identify the targets under strong lighting interference as well as FACTOR. This demonstrates that FACTOR has strong generalization capabilities and can identify interference from non-causal attributes. More visual comparison results are provided in Appendix.
\begin{figure}[t]
\begin{center}
\centerline{\includegraphics[width=0.45\textwidth]{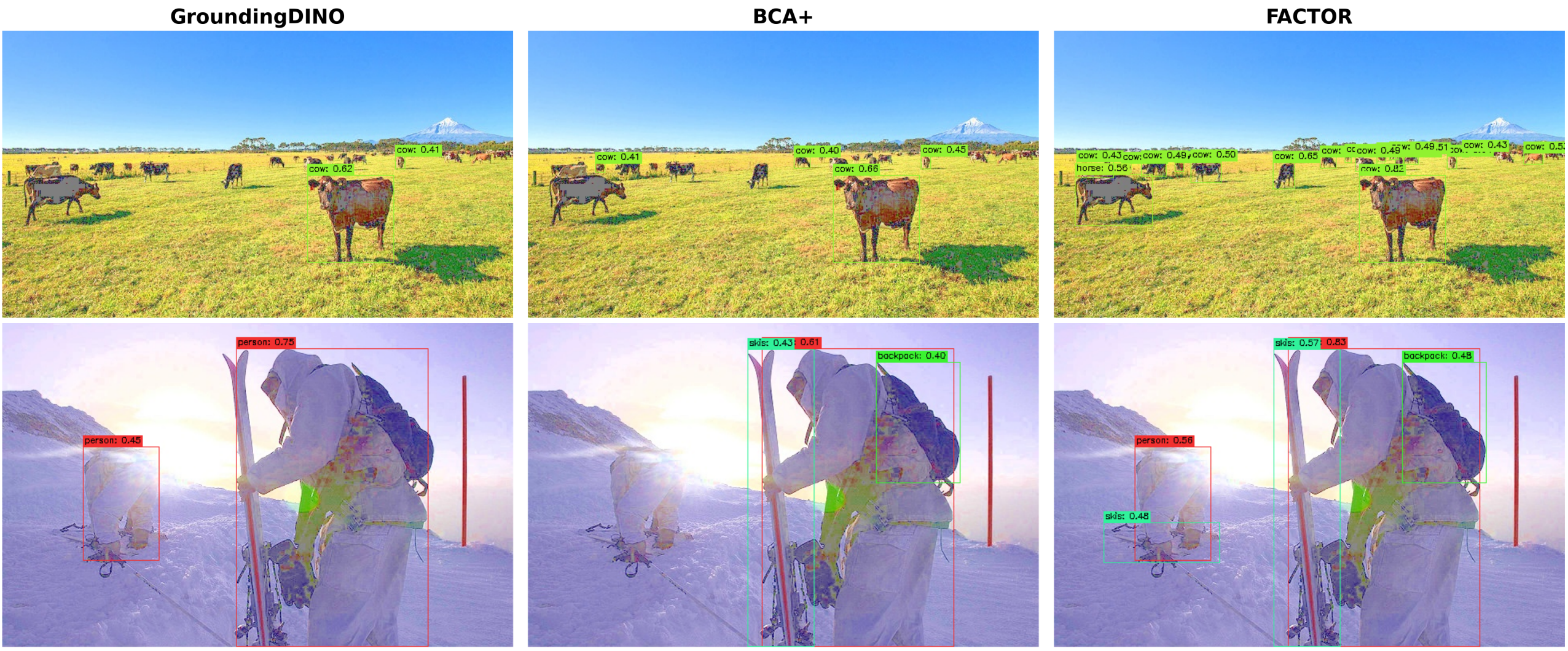}}
\caption{Visualization comparison among the baseline GroundingDINO, BCA+ and FACTOR. {\it Zoom in for best view.} }
\label{icml-historical}
\end{center}
\vskip -0.3in
\end{figure}

\subsection{Further Analysis}
\textbf{Ablation study.} Table~\ref{tab:Component} presents an ablation study to analyze the contribution of each component in FACTOR. Activating any single factor already leads to substantial improvements over the baseline, confirming that each component provides meaningful calibration signals.
Among single-factor variants, ASS achieves the largest gain, indicating that region-level attribute sensitivity plays a dominant role in identifying shift-affected regions, while ACR and CSS also provide consistent but weaker improvements.
Combining two factors further improves performance in most cases, with configurations involving ASS ((4) and (5)) approaching the full model, whereas the absence of ASS ((6)) leads to clear degradation.
The full FACTOR model achieves the best performance, demonstrating that jointly modeling region sensitivity, attribute–category relevance, and prediction discrepancy yields the most effective calibration.
\begin{table}[t]
\centering
\caption{Ablation study on key components of FACTOR. Results are
averaged over COCO-C. Backbone: Swin-T.}
\label{tab:Component}
\resizebox{0.3\textwidth}{!}{
\small
\begin{tabular}{c|c c c|c}
\hline
{Method} & ASS &  ACR & CSS  & $\text{mAP}_{50}$ \\ \hline
baseline & $\XSolidBrush$ & $\XSolidBrush$ & $\XSolidBrush$ & 21.61 \\ \hline
(1) & $\XSolidBrush$ & $\XSolidBrush$ & $\checkmark$ &  25.55\\
(2) & $\XSolidBrush$ & $\checkmark$ & $\XSolidBrush$ &   27.79\\
(3) & $\checkmark$ & $\XSolidBrush$ & $\XSolidBrush$ &  27.60 \\
\hline
(4) & $\checkmark$ & $\checkmark$ &  $\XSolidBrush$ &  27.72\\
(5) & $\checkmark$ & $\XSolidBrush$ & $\checkmark$ &  27.91\\
(6) & $\XSolidBrush$ & $\checkmark$ & $\checkmark$ &  26.18\\
\hline
FACTOR & $\checkmark$ & $\checkmark$ & $\checkmark$ & 29.30 \\
\hline
\end{tabular}}
\end{table}

\begin{figure}[t]
\begin{center}
\centerline{\includegraphics[width=0.9\columnwidth]{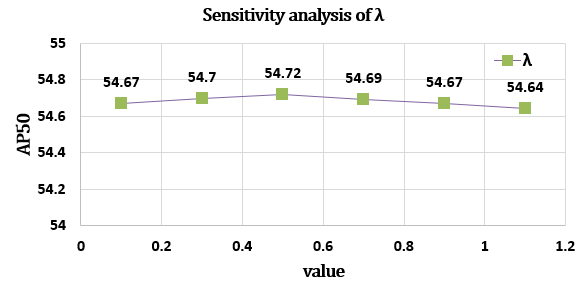}}
\caption{Hyperparameter sensitivity analysis on COCO-C (Swin-T backbone). FACTOR exhibits stable performance across a broad range.}
\label{fig:hyperparam_sensitivity}
\end{center}
\vskip -0.4 in
\end{figure}

\textbf{Hyperparameter sensitivity analysis.} The propose FACTOR  has only one parameter $\lambda$ in Eq.(\ref{eq:attribute_sensitivity}). We conduct a sensitivity analysis on the COCO-C Brit dataset, using the Grounding DINO-SwinT model. The value of $\lambda$ varies across a reasonable range $[0.1, 1.1]$. As shown in Fig~\ref{fig:hyperparam_sensitivity},  FACTOR achieves consistently strong performance across the tested range. In our method, $\lambda=0.5$. The stable performance across settings highlights the practicality of FACTOR under diverse test-time conditions.

\section{Conclusion}
We proposed FACTOR, a lightweight training-free TTA method for OVOD to address spurious correlation-induced performance degradation under domain shifts. By generating counterfactual images via attribute perturbations and fusing three complementary metrics—CSS, ASS and ACR. FACTOR explicitly models visual attributes and their causal links to predictions. Experiments confirm the effectiveness of our method. FACTOR not only enhances performance under diverse distribution shifts, but also opens a pathway toward more interpretable, efficient, and semantically grounded adaptation. Future work may extend FACTOR by integrating temporal memory for historical knowledge accumulation and introducing structured causal models to handle more complex environmental confounders in open-world deployment.


\section*{Impact Statement}

This work advances robust open-vocabulary object detection by providing an efficient test-time adaptation solution, enhancing VLM-based detector reliability in real-world scenarios, and supporting broader adoption in critical applications.

\nocite{belal2025vlod}
\nocite{carion2020end}
\nocite{chen2023stfar}
\nocite{cheng2024yolo}
\nocite{chenpali}
\nocite{cordts2016cityscapes}
\nocite{cui2022contrastive}
\nocite{devlin2019bert}
\nocite{everingham2015pascal}
\nocite{feng2023diverse}
\nocite{gu2022open}
\nocite{hansen2016cma}
\nocite{he2016deep}
\nocite{jaderberg2017population}
\nocite{iwasawa2021test}
\nocite{jia2021scaling}
\nocite{karmanov2024efficient}
\nocite{liang2025comprehensive}
\nocite{li2022grounded}
\nocite{li2025generalizing}
\nocite{liang2020we}
\nocite{liu2021swin}
\nocite{lin2014microsoft}
\nocite{michaelis2019benchmarking}
\nocite{liu2024grounding}
\nocite{minderer2022simpleopenvocabularyobjectdetection}
\nocite{pan2009survey}
\nocite{radford2021learning}
\nocite{real2017large}
\nocite{real2019regularized}
\nocite{ren2016faster}
\nocite{sakaridis2018semantic}
\nocite{sharifdeen2025tpt}
\nocite{sheng2025r}
\nocite{su2022revisiting}
\nocite{shlens2014noteskullbackleiblerdivergencelikelihood}
\nocite{shu2022test}
\nocite{vs2023towards}
\nocite{xue2015survey}
\nocite{wang2021tent}
\nocite{wang2025efficient}
\nocite{xu2021end}
\nocite{xue2015survey}
\nocite{yao2022detclip}
\nocite{yaofilip}
\nocite{yoo2024and}
\nocite{zareian2021open}
\nocite{zhai2022lit}
\nocite{zhang2024dual}
\nocite{zhang2024historical}
\nocite{zhangdino}
\nocite{zhou2022conditionalpromptlearningvisionlanguage}
\nocite{zhou2025bayesian}

\bibliography{example_paper}
\bibliographystyle{icml2026}


\newpage
\appendix
\section*{Appendix}
This appendix presents a detailed description of the counterfactual image generation process, including attribute modeling, intervention construction, and practical implementation choices used during test-time inference.
Furthermore, the appendix contains further analyses that deepen the understanding of FACTOR beyond the main text. These analyses include additional ablation results, factor interaction studies, and diagnostic evaluations that help explain the observed performance gains under diverse corruption settings. Finally, qualitative visual comparisons of detection results are provided to illustrate how counterfactual-guided calibration affects region-level predictions, highlighting its ability to suppress spurious attribute reliance while preserving semantically robust detections under domain shifts.

\section{More Related Work}
\subsection{Grounding DINO}
We adopt Grounding DINO as the base vision–language detector for open-vocabulary object detection. Given an input image $x$ and a set of category names $\mathcal{C} = \{c_1, \dots, c_C\}$, all category names are concatenated into a single text prompt, which is encoded by the text encoder of Grounding DINO. The detector then performs cross-modal reasoning between image features and text tokens to generate a fixed set of query-based region proposals.

For each proposal $r_i$, Grounding DINO outputs a normalized bounding box $b_i \in [0,1]^4$ and a category score vector $\mathbf{p}_i \in \mathbb{R}^C$. Category scores are obtained by matching query-level predictions with category-specific token spans in the text prompt, followed by a sigmoid activation. Proposals whose maximum score is below a confidence threshold are filtered out. Since the resulting category scores are not normalized, a log-softmax operation is applied to obtain a valid probability distribution over categories. Each remaining proposal is assigned a label $l_i = \arg\max_{c} p_i(c) $ with confidence score $\max_c p_i(c)$.
The final detection set is denoted as $\mathcal{R} = \{({f}_i, b_i, \mathbf{p}_i, l_i)\}_{i=1}^{M}$. All parameters of Grounding DINO are frozen during inference, and the detector is used purely as a black-box vision–language model to produce region-level predictions. Under distribution shifts, performance degradation mainly arises from semantic misalignment between the fixed text representations and shifted visual features, which motivates our test-time adaptation method.

\section{More Implementation Details.}
\subsection{Algorithm}
Algorithm~\ref{alg:cftta} summarizes FACTOR, a training-free test-time adaptation framework for open-vocabulary object detection.
Given a test image, FACTOR first estimates counterfactual prediction shifts by comparing the detector outputs on the original image and a composed attribute-level counterfactual.
It then performs attribute-guided logit calibration and ROI-level score refinement to suppress categories and regions that are unstable under counterfactual perturbations.
The entire procedure requires only a single additional forward pass and introduces no training-time overhead.

\begin{algorithm}[t]
\caption{FACTOR: Counterfactual Training-Free Test-Time Adaptation}
\label{alg:cftta}
\begin{algorithmic}[1]
\STATE {\bfseries Input:} Test image $x$; attribute set $\mathcal{A}$; category set $\mathcal{C}$; pretrained detector $\mathcal{M}$; bert encoder $\mathcal{B}$; hyperparameter $\lambda$
\STATE {\bfseries Output:} Adapted detections $\{(r_i,\hat{y}_i,s_i')\}_{i=1}^{N}$

\vspace{0.4em}
\STATE {\bfseries Stage I:  Counterfactual Probing}
\STATE $\{r_i,\mathbf{f}_i,\mathbf{z}_i,\mathbf{p}_i\}_{i=1}^{N} \leftarrow \mathcal{M}(x)$
\STATE $x^{cf} \leftarrow \textsc{ComposeAttributes}(x,\mathcal{\mathcal{A}})$
\STATE $\{\tilde{\mathbf{p}}_i\}_{i=1}^{N} \leftarrow \textsc{MatchAndPredict}(\mathcal{M}(x^{cf}),\{r_i\})$
\STATE $\mathbf{T}^{\mathrm{attr}} \leftarrow \mathcal{B}(\mathcal{A})$
\STATE $\mathbf{T}^{\mathrm{cls}} \leftarrow \mathcal{B}(\mathcal{C})$
\vspace{0.4em}
\STATE {\bfseries Stage II: Invariance-Guided Calibration}
\STATE $\mu \leftarrow \frac{1}{N}\sum_{i}\sum_{c} p_i(c)\log\frac{p_i(c)}{\tilde{p}_i(c)}$
\FOR{each region $r_i$} 
    \STATE $\mathrm{ASS}_{i,a}\leftarrow\sigma\left(\mathbf{f}_i^{\top}\mathbf{T}^{\mathrm{attr}}_a\right)$ 
    \STATE $\mathrm{ACR}_{a,c}\leftarrow\sigma\!\left((\mathbf{T}^{\mathrm{attr}}_a)^{\top}\mathbf{T}^{\mathrm{cls}}_c\right)$
    \STATE $\mathrm{KL}_i \leftarrow \sum_{c} p_i(c)\log\frac{p_i(c)}{\tilde{p}_i(c)}$
    \STATE $\mathrm{CSS}_i \leftarrow \sigma(\mathrm{KL}_i-\mu)$
    \STATE $\Delta_i(c) \leftarrow \frac{1}{|\mathcal{A}|}\sum_{a\in\mathcal{A}}
    \mathrm{ACR}_{a,c}\,
    \mathrm{ASS}_{i,a}\,
    \mathrm{CSS}_i$
    \STATE $\mathbf{logits}_i' \leftarrow \mathbf{logits}_i - \lambda \boldsymbol{\Delta}_i$
    \STATE $\mathbf{p}_i' \leftarrow \sigma(\mathbf{logits}_i')$
    \STATE $\hat{y}_i \leftarrow \arg\max_c p_i'(c)$
    \STATE $\bar{\Delta}_i \leftarrow \sum_c p_i'(c)\Delta_i(c)$
    \STATE $s_i' \leftarrow s_i \cdot \exp(-\bar{\Delta}_i)$
\ENDFOR

\STATE {\bfseries return} $\{(r_i,\hat{y}_i,s_i')\}_{i=1}^{N}$
\end{algorithmic}
\end{algorithm}

\subsection{Counterfactual Operations}
This subsection details the concrete attribute-level transformations used to construct counterfactual images.
Each operation intervenes on a specific non-causal appearance factor while preserving object identity, spatial structure, and high-level semantics.
All transformations are deterministic, applied channel-wise, and parameterized with fixed intensities to ensure reproducibility and semantic consistency.

\textbf{Illumination (Brightness).} We simulate brightness variation via monotonic gamma correction:
\begin{equation}
\mathcal{T}_{\text{brit}}(x) = x^{\gamma'}, \quad \gamma' > 1,
\end{equation}
which uniformly darkens the image and suppresses over-exposed regions. This operation perturbs global luminance statistics without altering color relationships or geometric content. In practice, the effective exponent $\gamma'$ is implemented through an inverse parameterization, ensuring a consistent darkening effect rather than artificial enhancement.

\textbf{Contrast.} Global contrast variation is modeled by linear intensity scaling:
\begin{equation}
\mathcal{T}_{\text{cont}}(x) = \alpha x, \quad \alpha \in (0,1),
\end{equation}
which reduces the dynamic range of pixel intensities.
This operation degrades edge sharpness and local discriminability while preserving object boundaries and relative color composition.

\textbf{Blur.} We apply Gaussian blurring to suppress high-frequency visual cues:
\begin{equation}
\mathcal{T}_{\text{blur}}(x) = G_k * x,
\end{equation}
where $G_k$ denotes a Gaussian kernel of size $k$. Blurring selectively removes fine-grained texture and edge details while retaining coarse object shape and spatial layout.

\textbf{Noise.} Additive Gaussian noise is injected to perturb pixel-level statistics:
\begin{equation}
\mathcal{T}_{\text{noise}}(x) = x + \varepsilon, \quad \varepsilon \sim \mathcal{N}(0, \sigma^2).
\end{equation}
This operation simulates sensor noise and compression artifacts commonly encountered in real-world data acquisition, without introducing structured distortions.

\textbf{Texture.} Texture degradation is simulated via resolution perturbation:
\begin{equation}
\mathcal{T}_{\text{text}}(x) = \mathrm{Resize}_{\uparrow}(\mathrm{Resize}_{\downarrow}(x, \theta)).
\end{equation}
where $\theta$ denotes the sampling density. Downsampling followed by upsampling disrupts fine-grained texture patterns while preserving global object shape and spatial alignment, effectively reducing reliance on local texture cues.

\textbf{Weather.}
Weather-related appearance shifts (e.g., fog or haze) are approximated through linear blending with a uniform intensity field:
\begin{equation}
\mathcal{T}_{\text{weath}}(x) = (1 - \beta)x + \beta \mathbf{1}.
\end{equation}
This operation attenuates contrast and visibility in a spatially uniform manner, mimicking atmospheric scattering effects without occluding objects or altering scene geometry.  

All parameters $(\gamma'=0.3, \alpha=0.9, k=3, \sigma=2, \beta=0.1)$ are fixed across all images and datasets to induce perceptible yet semantically invariant perturbations. Crucially, none of the above operations modify object identity, spatial arrangement, or category-level semantics. As a result, any prediction discrepancy between the original image $x$ and its counterfactual counterpart $x^{\mathrm{cf}}$ can be attributed to the detector’s sensitivity to attribute-level appearance shifts, rather than to content corruption or semantic alteration.

\begin{figure}[t]

\begin{center}
\centerline{\includegraphics[width=\columnwidth]{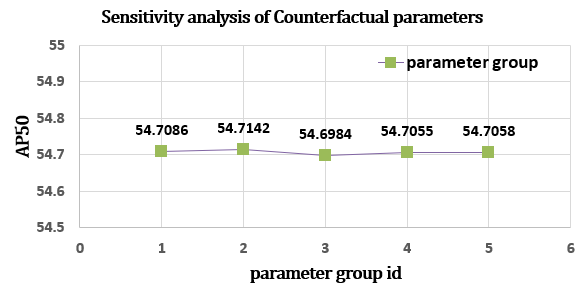}}
\caption{Counterfactual image hyperparameter sensitivity analysis on COCO-C (Swin-T backbone). FACTOR exhibits stable performance across a broad range of shift scenarios.}
\label{Counterfactual_parameter}
\end{center}
\vskip -0.2in
\end{figure}

{\bf Parameter analysis.} To verify the impact of the above mentioned parameters, we conduct five comparative experiments on COCO-C Brit (SwinT). The specific experimental settings are as follows $(\gamma', \alpha, k, \sigma, \theta ,\beta)$: (1) (0.9,0.98,1,0.5,0.98,0.03); (2) 0.3, 0.90, 3, 2, 0.95, 0.10; (3) (0.15, 0.75, 5, 5, 0.85, 0.25); (4) (1.0, 1.0, 5, 4, 0.85, 0.0); (5) (0.25, 0.85, 3, 1, 1.0, 0.20). The parameters have the following specific meanings: (1) extremely weak perturbation strength; (2) balanced strength; (3) strong attribute interference; (4) isolation of low-level visual attributes such as texture and blur; (5) simulation of real-world distribution shift scenarios. As shown in Fig.~\ref{Counterfactual_parameter}, the model can obtain robust results in various simulated scenarios, demonstrating its ability to adjust the correction intensity according to different scenarios. 

\section{More Experimental Results.}

\textbf{Full results on ablation study.} In Table~\ref{tab:ablationall} , we report the contribution of each component across all 15 corruption types on COCO-C, as well as the average AP50. Backbone is Swin-T. Activating any single factor already leads to substantial improvements over the baseline. The full FACTOR model achieves the best performance, demonstrating that jointly modeling region sensitivity, attribute–category relevance, and prediction variance yields the most effective and stable calibration.

\begin{table*}[t]
\centering
\caption{Object detection performance comparison with the state-of-the-art methods on COCO-C.
Metric: mean Average Precision@50(\%)}
\label{tab:ablationall}
\resizebox{\textwidth}{!}{
\begin{tabular}{c|c c c|c c c c c c c c c c c c c c c|c}
\hline
{Methods}  & ASS &  ACR & CSS& Brit & Contr & Defoc & Elast & Fog & Frost & Gauss & Glass & Impul & Jpeg & Motn & Pixel & Shot & Snow & Zoom & Average \\\hline

 \hline
GDINO & $\XSolidBrush$ & $\XSolidBrush$ & $\XSolidBrush$  &42.91&23.83&19.45&23.76&43.26&30.36&15.03&10.22&15.87&23.77&16.56&7.31&16.99&25.51&9.35& 21.61 \\
 \hline
(1) & $\XSolidBrush$ & $\XSolidBrush$ & $\checkmark$ & 47.54&29.27&23.67&26.54&49.94&34.98&19.33&12.97&20.51&26.79&19.90&11.01&20.78&29.05&10.95& 25.55\\
(2) & $\XSolidBrush$ & $\checkmark$ & $\XSolidBrush$ & 52.36&30.90&24.91&30.44&52.95&38.78&20.42&13.92&21.85&30.11&21.17&11.85&22.72&32.81&11.65&  27.79\\
(3) & $\checkmark$ & $\XSolidBrush$ & $\XSolidBrush$ &50.84&30.76&25.00&31.03&51.80&38.82&20.53&14.10&21.95&29.62&21.11&12.05&22.54&32.58&11.28&27.60 \\
\hline
(4) & $\checkmark$ & $\checkmark$ &  $\XSolidBrush$ & 51.13&30.88&25.08&31.13&52.05&39.00&20.61&14.15&22.04&29.81&21.17&12.12&22.64&32.74&11.32&27.72\\
(5) & $\checkmark$ & $\XSolidBrush$ & $\checkmark$ & 51.47&31.21&25.12&31.24&52.69&39.22&20.90&14.19&22.27&29.85&21.24&12.22&22.82&32.86&11.41&27.91\\
(6) & $\XSolidBrush$ & $\checkmark$ & $\checkmark$ & 48.83&29.75&24.13&27.33&50.97&36.04&19.68&13.21&20.94&27.69&20.29&11.24&21.34&30.08&11.13& 26.18\\
\hline
FACTOR & $\checkmark$ & $\checkmark$ & $\checkmark$ & 54.71 & 32.15 & 25.99 & 32.69 & 55.09 & 41.04 & 21.76 & 14.76 & 23.33 & 31.84 & 22.22 & 12.87 & 23.98 & 34.98 &12.05 & 29.30 \\
\hline
\end{tabular}}
\vspace{-0.5cm}
\end{table*}

\textbf{Number of Attributes.} To assess how the size of the attribute set in counterfactual processing influences model performance, we conduct a sensitivity analysis on the COCO-C Brit dataset using the Grounding DINO-SwinT model. We construct incremental attribute subsets: starting with 1 attribute (brightness), expanding to 2 attributes (adding contrast), then 4 attributes (including noise and blur), and finally using all 6 attributes (the full set in our method).From Table \ref{tab:NumberofAttributes}, we observe a monotonic improvement in $\text{mAP}_{50}$ as more attributes are incorporated. This trend demonstrates that a richer attribute set enables the model to capture more diverse attribute-induced distribution shifts, thereby enhancing robustness to corruptions on COCO-C Brit. 
Notably, the marginal performance gain diminishes with larger attribute sets (e.g., only 1.00 point from 4 to 6 attributes), indicating that the 6-attribute set already covers the key shift factors relevant to this dataset. This result validates the design of our full attribute set while highlighting that even partial subsets can yield meaningful performance improvements. Although more attributes may bring performance improvements, considering the computational complexity, our work chooses a number of attributes of 6.

\begin{table*}[t]
\centering
\vspace{0.5cm}
\caption{FACTOR performance with varying numbers of counterfactual attributes on COCO-C Brit.}
\label{tab:NumberofAttributes}
\resizebox{0.7\textwidth}{!}{
\begin{tabular}{c|c c c c c c|c }
\hline
Number of Attribute & brightness &contrast&noise & blur& texture&weather&$\text{mAP}_{50}$\\ \hline
1&$\checkmark$& $\XSolidBrush$ & $\XSolidBrush$ & $\XSolidBrush$ & $\XSolidBrush$& $\XSolidBrush$ &  48.10  \\
2&$\checkmark$& $\checkmark$& $\XSolidBrush$ & $\XSolidBrush$ & $\XSolidBrush$& $\XSolidBrush$ & 52.01  \\
4&$\checkmark$& $\checkmark$&$\checkmark$&$\checkmark$ & $\XSolidBrush$& $\XSolidBrush$ & 53.71  \\
6&$\checkmark$& $\checkmark$&$\checkmark$&$\checkmark$&$\checkmark$&$\checkmark$& 54.71  \\
\hline
\end{tabular}}
\vspace{-0.5cm}
\end{table*}

\begin{table}[t]
\centering
\caption{Statistical significance (p-values) of performance differences between FACTOR and BCA+ on the COCO-C dataset with Swin-T.}
\label{tab:Statistical}
\resizebox{0.45\textwidth}{!}{
\begin{tabular}{c|c c|c}
\hline
Metric & FACTOR (Mean±Std) & BCA+ (Mean±Std) & p-value \\\hline
AP50  &29.29±0&25.06±0 & 6.533246e-20\\
\hline
\end{tabular}}
\vspace{-0.5cm}
\end{table}

\textbf{Statistical verification.} To further validate the performance advantage of the proposed FACTOR framework, we conducted a statistical significance analysis against BCA+ using the Wilcoxon signed-rank test, as summarized in Table \ref{tab:Statistical}. We select BCA+ for comparison because it represents the state-of-the-art in training-free test-time adaptation for open-vocabulary object detection, providing a strong and fair baseline. Since FACTOR incorporates stochastic elements in its counterfactual generation process (e.g., attribute perturbations with random parameters) while BCA+ is deterministic, we designed the statistical analysis as follows:
We executed FACTOR with 10 different random seeds on the COCO-C dataset with Swin-T backbone, collecting performance metrics for average of the 15 corruption types. For BCA+, we report its deterministic results with zero variance. The Wilcoxon signed-rank test is a non-parametric method suitable for paired comparisons without assuming data normality, making it appropriate for evaluating detection performance across multiple experimental runs. In our case, we employed a one-sample Wilcoxon test comparing FACTOR's multi-seed distribution against BCA+'s fixed value for average AP50.
As shown in Table~\ref{tab:Statistical}, FACTOR demonstrates statistically significant improvements over BCA+ across average AP50. The corresponding p-values are all well below the 0.05 significance threshold, indicating that the performance improvement of FACTOR compared to BCA+ is statistically significant and not due to random fluctuations.

\begin{figure}[t]
\begin{center}
\centerline{\includegraphics[width=\linewidth]{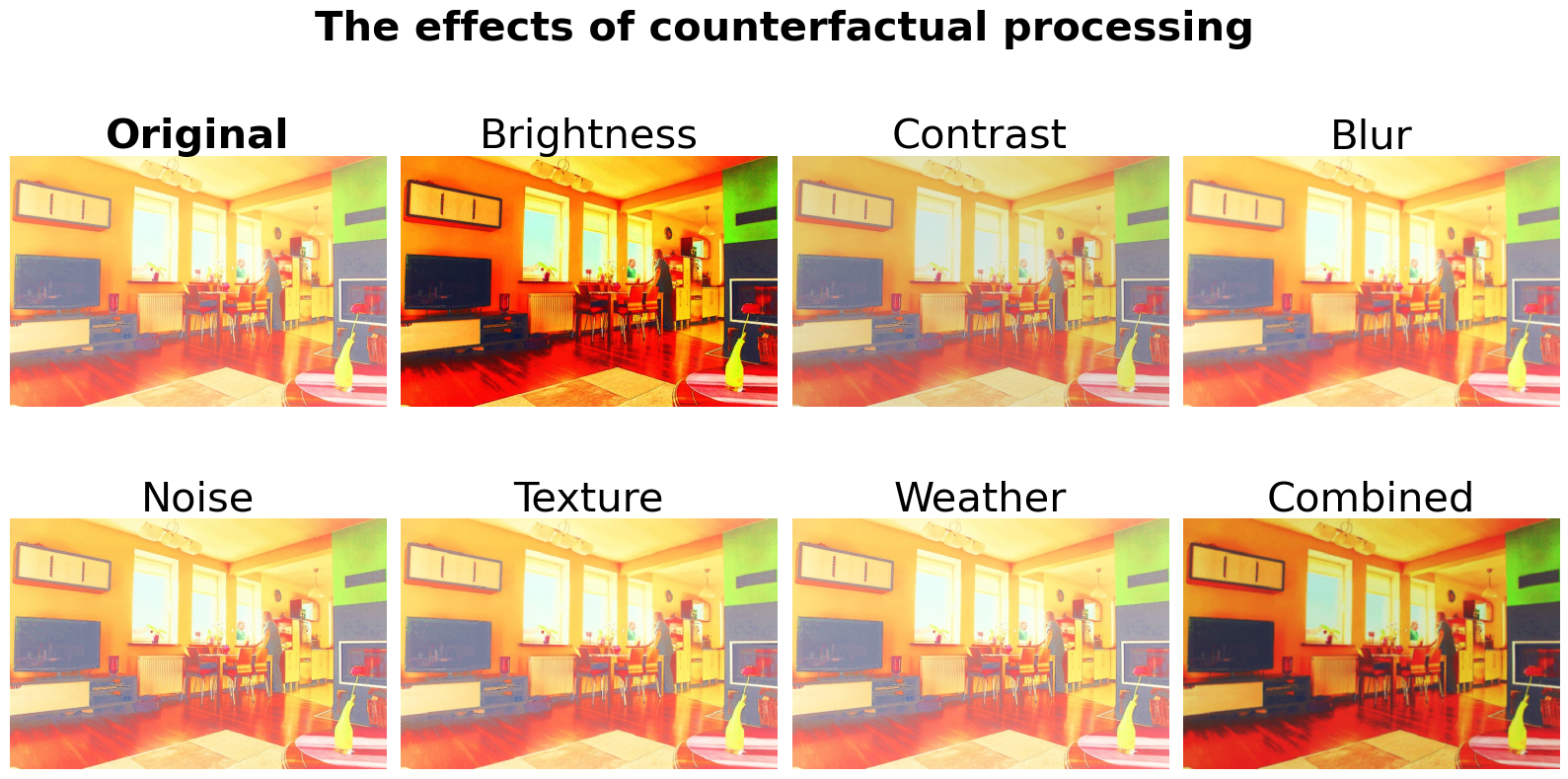}}
\caption{The effect of counterfactual processing on a sample image. All effects are based on the original image. {\it Zoom in for best view.}}
\label{fig:attributes_look}
\end{center}
\vskip -0.2in
\end{figure}

\textbf{Effects after counterfactual processing.} To illustrate the visual and quantitative impacts of our counterfactual attribute transformations, we present both a qualitative visualization (Fig.\ref{fig:attributes_look}) and quantitative metrics (Table \ref{tab:attributes_look}) for each transformation. 
To quantify the discrepancy between the original image $x_{\text{orig}}$ and the counterfactual image $x_{\text{cf}}$, we introduce three core metrics to measure pixel-level perturbations. 
First, the average pixel difference $\Delta\mu$ is defined as the mean of absolute pixel-wise differences across all spatial locations and RGB channels:
\begin{equation}
\Delta\mu = \frac{1}{N} \sum_{i,j,c} \left| I_{\text{orig}}(i,j,c) - I_{\text{cf}}(i,j,c) \right|,
\label{eq:avg_pixel_diff}
\end{equation}
where $N = H \times W \times 3$ denotes the total number of pixels (height $H$, width $W$, and three RGB channels $c$), and the $\pm$ notation in subsequent results represents the standard deviation of pixel differences. 
Second, we compute the maximum pixel difference $\Delta_{\text{max}}$ to capture the extreme perturbation on individual pixels:
\begin{equation}
\Delta_{\text{max}} = \max_{i,j,c} \left| I_{\text{orig}}(i,j,c) - I_{\text{cf}}(i,j,c) \right|,
\label{eq:max_pixel_diff}
\end{equation}
which reflects the largest deviation between the original and counterfactual images at any pixel location. 
Finally, to normalize the perturbation strength to an interpretable scale, we calculate the relative change (RC) as:
\begin{equation}
\text{RC (\%)} = \frac{\Delta\mu}{255} \times 100,
\label{eq:relative_change}
\end{equation}
where 255 is the maximum value of an 8-bit pixel, ensuring the relative change is expressed as a percentage to facilitate cross-transformation comparison.

\begin{table}[h]
\centering
\caption{Quantitative metrics for counterfactual transformations. All values are computed against the original image. The ``Combined'' transformation exhibits the strongest overall perturbation (highest $\Delta\mu$ and relative change).}
\label{tab:attributes_look}
\resizebox{0.4\textwidth}{!}{
\begin{tabular}{l|c c c}
\hline
Attribute &  $\Delta\mu$  & $\Delta_{\text{max}}$ & RC (\%)\\ \hline
Original  \\\hline
Brightness & $18.67 \pm 14.39$ & $37.00$ & $7.32$ \\
Contrast & $17.97 \pm 6.08$ & $25.00$ & $7.05$ \\
Blur & $3.02 \pm 5.53$ & $111.00$ & $1.18$ \\
Noise & $1.54 \pm 1.31$ & $10.00$ & $0.60$ \\
Texture & $4.07 \pm 8.74$ & $180.00$ & $1.59$ \\
Weather & $7.07 \pm 6.09$ & $25.00$ & $2.77$ \\
\hline
Combined & $22.55 \pm 11.99$ & $142.00$ & $8.84$ \\
\hline
\end{tabular}}
\end{table}

\begin{figure*}[!t]
    \centering
    
    \begin{subfigure}{\textwidth}
        \centering
        \includegraphics[width=\textwidth]{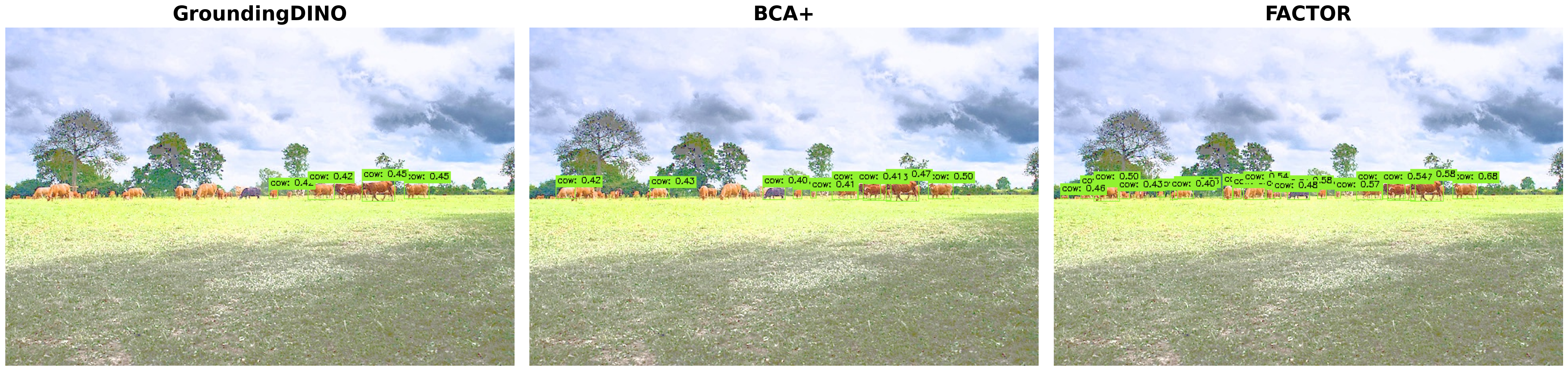}
        \caption{Dense detection scenarios involving small objects.}
        \label{comparison_1}
    \end{subfigure}
    
    \vspace{0.8em} 

    \begin{subfigure}{\textwidth}
        \centering
        \includegraphics[width=\textwidth]{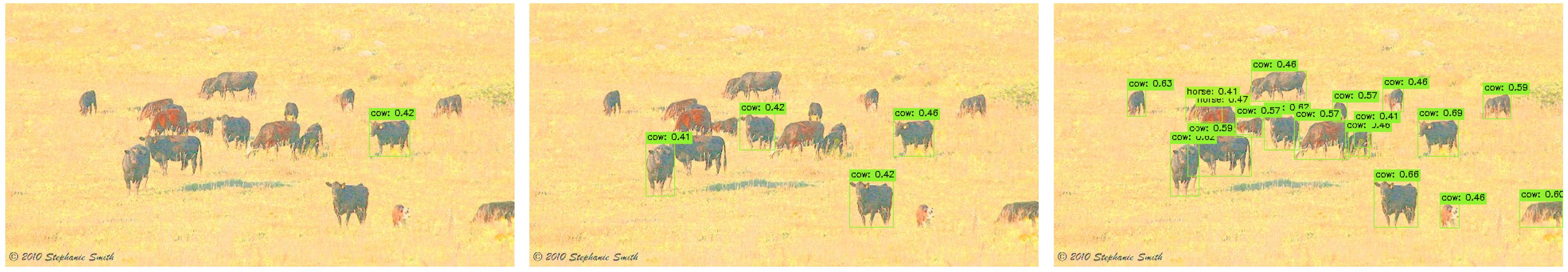}
        \caption{Dense detection scenarios involving small objects.}
        \label{comparison_2}
    \end{subfigure}

    \vspace{0.8em}

    \begin{subfigure}{\textwidth}
        \centering
        \includegraphics[width=\textwidth]{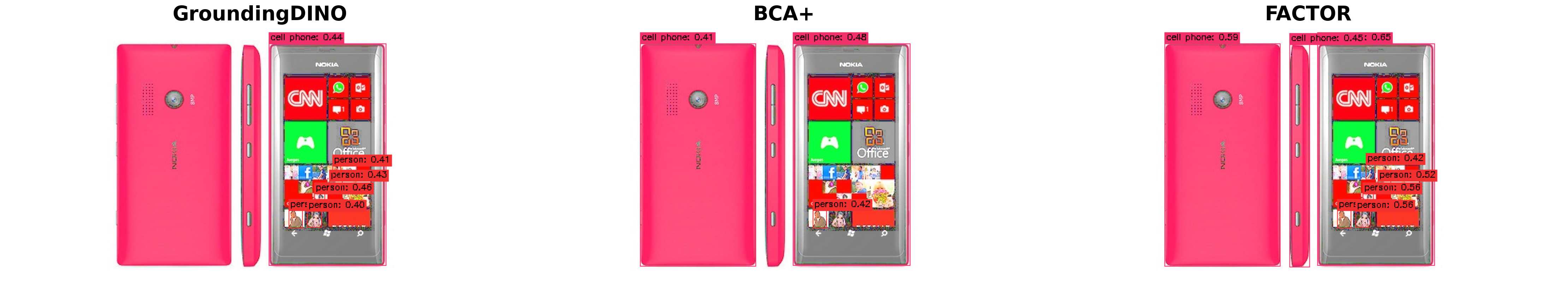}
        \caption{Single-category detection scenarios with diverse features.}
        \label{comparison_3}
    \end{subfigure}

    \vspace{0.8em}

    \begin{subfigure}{\textwidth}
        \centering
        \includegraphics[width=\textwidth]{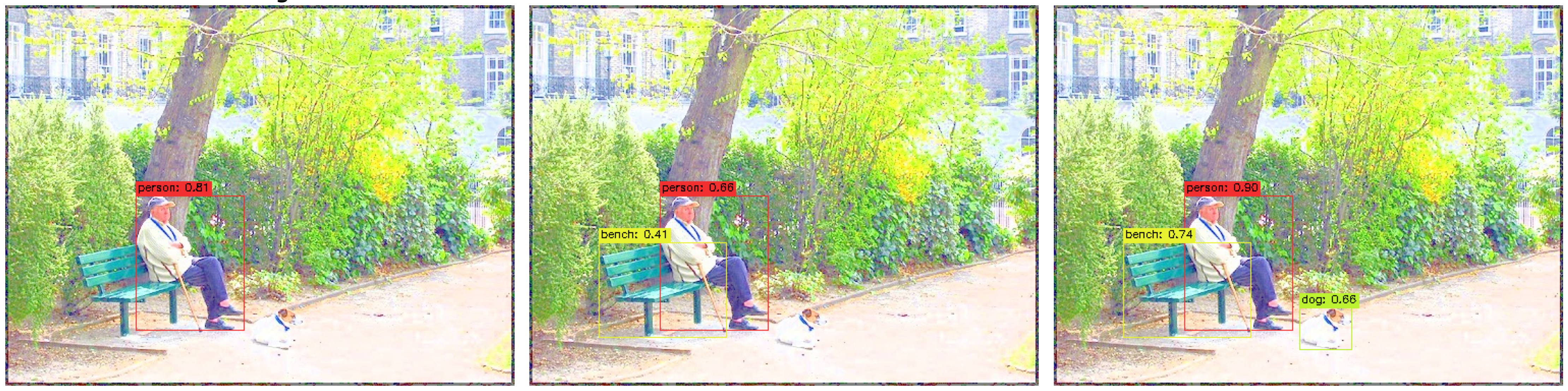}
        \caption{Detection scenarios across multiple categories.}
        \label{comparison_5}
    \end{subfigure}

    \vspace{0.8em}

    \begin{subfigure}{\textwidth}
        \centering
        \includegraphics[width=\textwidth]{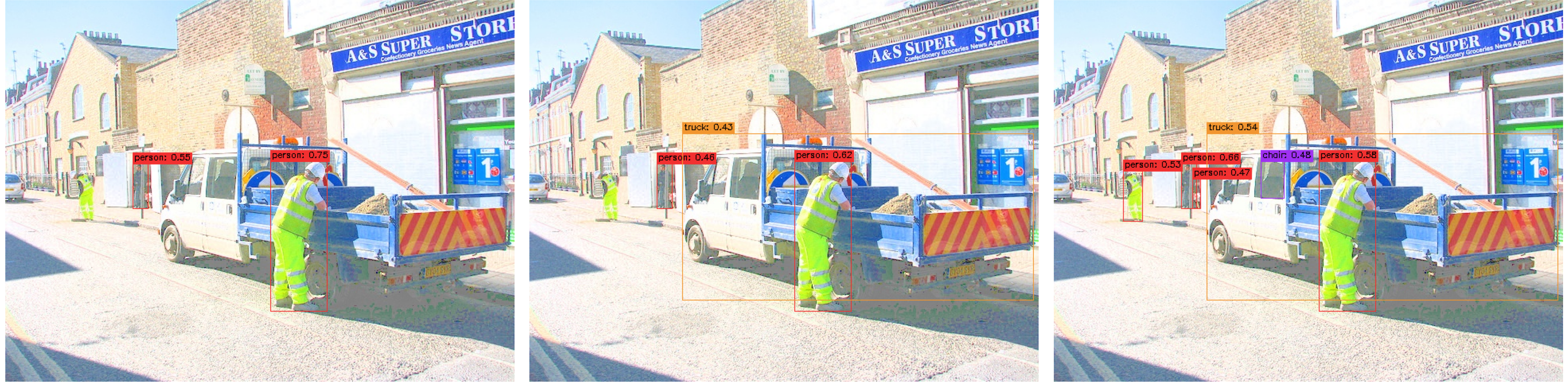}
        \caption{Detection scenarios across multiple categories.}
        \label{comparison_7}
    \end{subfigure}

    \caption{Visual comparison results on COCO-C and SwinT across various challenging scenarios. {\it Zoom in for best view.}}
    \label{fig:all_comparisons}
\end{figure*}

\textbf{Comparison of detection results.} In order to more intuitively understand the superiority of FACTOR, we conducted a visual comparison experiment between FACTOR, Grounding DINO, and BCA+, on the COCO-C Brit dataset using the Grounding DINO-SwinT model, and selected some images for intuitive analysis. From Fig.~\ref{comparison_1}-\ref{comparison_7}, we can observe that FACTOR demonstrates a clear advantage in both anchor box scores and the number of detected anchor boxes. This proves the effectiveness of FACTOR. From Fig.~\ref{comparison_1} and \ref{comparison_2}, we can see that FACTOR exhibits extremely high performance in detecting dense objects and relatively small targets. In Fig.~\ref{comparison_3}, FACTOR successfully identifies a side-facing phone that neither Grounding DINO nor BCA+ could detect. In Figs.~\ref{comparison_1} and \ref{comparison_7}, FACTOR identifies objects of other categories that are missed by other models.

\section{Limitations and Future Works}
While FACTOR demonstrates consistent improvements for test-time adaptation in open-vocabulary object detection, several limitations remain. First, the effectiveness of counterfactual-guided calibration depends on the quality of the constructed counterfactual views. Although attribute perturbations are designed to target spurious correlations, imperfect attribute modeling or incomplete coverage of nuisance factors may limit the ability to fully expose prediction instability, especially under complex real-world corruptions. Future work could explore more expressive attribute representations or adaptive intervention strategies that better capture diverse domain-specific biases.

Second, FACTOR performs calibration based on region-level sensitivity signals derived from paired predictions. While this design avoids model updates and preserves efficiency, it does not explicitly reason about temporal or cross-image consistency. Extending the framework to leverage multi-image or video-level context at test time may further improve accuracy, particularly in scenarios with strong temporal coherence.

Finally, although FACTOR is training-free, it still incurs additional computation due to counterfactual forward passes. Future research may investigate lightweight approximation strategies or selective intervention mechanisms that reduce overhead while maintaining calibration effectiveness. Beyond object detection, we believe the counterfactual-guided sensitivity calibration paradigm can be generalized to other vision–language tasks, offering a promising direction for robust test-time adaptation under distribution shifts.

\end{document}